\documentclass[a4paper,fleqn]{cas-sc}

\usepackage[authoryear,longnamesfirst]{natbib}

\def\tsc#1{\csdef{#1}{\textsc{\lowercase{#1}}\xspace}}
\tsc{WGM}
\tsc{QE}

\begin{document}
\let\WriteBookmarks\relax
\def\floatpagepagefraction{1}
\def\textpagefraction{.001}

\shorttitle{Physics-Informed Error Field Learning}    

\shortauthors{J.Y. Sun and Y. Zhang}  

\title [mode = title]{Physics-Informed Error Field Learning: A Post-Training Optimization Framework for Physics-Informed Neural Networks}  



%

\author[1]{Jiuyun Sun}
\affiliation[1]{organization={College of Mathematics and Systems Science, Shandong University of Science and Technology},
	addressline={Qianwangang Road 579}, 
	city={Qingdao},
	postcode={266590}, 
	state={Shandong},
	country={China}}
\author[1]{Yong Zhang}







\begin{abstract}
Physics-Informed Neural Networks (PINNs) have emerged as an important class of numerical methods for solving partial differential equations (PDEs). However, during the late-stage optimization process, further parameter updates often yield diminishing accuracy improvements while increasing computational costs. To address this issue, this paper proposes a Physics-Informed Error Field Learning (PIEFL) framework for PINNs. Unlike conventional approaches that continuously approximate the solution field using a single network, PIEFL introduces an auxiliary error network after the primary network achieves satisfactory accuracy and shifts the learning objective from the solution field to the error field. By deriving error control equations under physical constraints, the error network learns the discrepancy between the current approximation and the exact solution, and the learned error correction is combined with the primary prediction to improve solution accuracy.
The proposed framework avoids continuous optimization of the entire solution space and focuses computational resources on correcting existing prediction errors. Moreover, PIEFL requires no modification to the primary network architecture, making it compatible with existing PINN models and applicable as a general post-training optimization strategy. Numerical experiments on representative PDEs demonstrate that PIEFL achieves higher solution accuracy under the same computational budget, validating its effectiveness in improving the performance of PINNs.
\end{abstract}




\maketitle

\section{Introduction}\label{}

Partial differential equations (PDEs) serve as fundamental mathematical tools for describing complex physical processes in the natural sciences and engineering, with broad applications in fields including fluid mechanics, materials science, environmental science, and biomedicine \cite{evans2022partial, cai2021flow, pun2019physically, cleary2004discrete, sobie2011systems}. Developing efficient and accurate approaches for solving PDEs has long been a central research topic in scientific computing. In recent years, Physics-Informed Neural Networks (PINNs) have emerged as a novel computational framework that integrates data-driven learning with physical constraints, providing a new paradigm for solving PDEs \cite{raissi2019physics}. By incorporating physical residuals into the loss function of neural networks, PINNs eliminate the complex meshing procedures required by conventional numerical methods and have demonstrated considerable potential in applications such as fluid dynamics, solid mechanics, and biomedical engineering \cite{cai2021physics, hu2024physics, roquemen2025recent, rao2021physics}.

Although PINNs have demonstrated significant potential for solving PDEs, they still face substantial challenges in terms of computational efficiency, training stability, and the capability to achieve high-precision solutions \cite{cuomo2022scientific, karniadakis2021physics, krishnapriyan2021characterizing}. To address the optimization difficulties, slow convergence, and limited approximation capability for complex physical fields encountered during PINN training, existing studies have primarily focused on enhancing the representation capacity of solution networks and improving the training strategies \cite{jagtap2020adaptive, yang2021b, lin2022two}.
On the one hand, more effective neural network architectures have been developed, including residual connections, Fourier features, and other advanced feature mapping techniques, to improve the capability of neural networks in representing complex physical fields \cite{wang2021understanding, wang2021eigenvector, song2023simulating, gao2021phygeonet, ren2022phycrnet, sun2024physical}. On the other hand, strategies such as adaptive sampling and domain decomposition have been introduced to optimize the distribution of training points, enabling the network to concentrate more effectively on regions where physical constraints are difficult to satisfy \cite{wu2023comprehensive, tang2023pinns, jagtap2020conservative, jagtap2020extended, moseley2023finite}. Furthermore, adaptive loss weighting, gradient balancing, and advanced optimization algorithms have been employed to alleviate the imbalance among different loss components and improve the stability and convergence behavior of the training process \cite{wang2021understanding, xiang2022self, feng2026adaptive, li2022dynamic, hou2023enhancing, wang2025famaw, tao2025lstm}.
These approaches enhance the performance of PINNs from different perspectives. However, they share a common characteristic: they fundamentally rely on continuously optimizing the parameters of the solution network to approximate the exact solution.

However, an issue that has not yet received sufficient attention is that, during the later stages of PINN training, further optimization of network parameters often results in diminishing improvements in solution accuracy, while the computational cost continues to increase, leading to a pronounced diminishing-return effect. At this stage, the solution network has typically captured the overall structure of the target solution field, and a substantial amount of computational resources is devoted to further refining the network parameters, yet only marginal improvements in the solution error can be achieved. Therefore, an important question naturally arises: Is it still necessary to continuously optimize the solution network, or can higher solution accuracy be achieved more efficiently by shifting the learning objective?

Based on the above observations, this paper proposes a novel Physics-Informed Error Field Learning (PIEFL) framework, which provides a new strategy for the late-stage optimization of PINNs. Unlike existing methods that continuously learn the solution field, the proposed approach first trains a PINN to obtain an accurate approximation of the solution field, which is referred to as the primary network. Once the primary network reaches a satisfactory level of accuracy, the learning objective is shifted from the solution field to the error field.
By formulating error control equations, the prediction errors of the primary network are transformed into an error field that satisfies the underlying physical constraints, and an auxiliary PINN is employed to learn this field. Finally, the learned error field is superimposed onto the prediction of the primary network to obtain a more accurate numerical solution.

Compared with continuously optimizing a solution network that is already approaching convergence, PIEFL introduces an independent error network to directly learn the residual errors in the existing prediction, thereby achieving further accuracy improvements with only marginal additional computational overhead. Therefore, this work focuses on improving the computational efficiency of PINNs during the late-stage optimization process. Specifically, it achieves more efficient error correction by shifting the learning objective while preserving the original physics-constrained solution framework.

The proposed PIEFL framework does not require any modification to the fundamental architecture of existing PINNs and serves as a general-purpose post-optimization strategy that is compatible with existing enhancement techniques, including network architecture improvements, adaptive sampling, and loss weighting strategies. The main contributions of this work are summarized as follows:

\begin{itemize}
	\item A Physics-Informed Error Field Learning (PIEFL) framework is proposed. Unlike conventional PINNs that continuously optimize the solution network, PIEFL treats the error field as an independent learning target and shifts the optimization objective from solution field learning to error field learning.
	
	\item The corresponding physics-based control equations for the error field are established. By deriving the physical constraints satisfied by the error field, the error learning process is incorporated into the PINNs framework, ensuring that the error network remains subject to the same physical constraints as the original system.
	
	\item An efficient late-stage optimization strategy for PINNs is developed. By switching from solution optimization to error field learning during the diminishing-return stage, PIEFL further reduces prediction errors with lower computational costs.
	
	\item The effectiveness of PIEFL is validated through several representative PDE problems. Numerical results demonstrate that, compared with continuous training strategies, PIEFL achieves higher solution accuracy with lower computational costs, while exhibiting strong applicability and compatibility.
\end{itemize}

The remainder of this paper is organized as follows. Section 2 presents the theoretical foundation and implementation methodology of PIEFL in detail. Section 3 describes the experimental setup, analyzes the numerical results, and provides comparative validations of the proposed framework. Section 4 concludes this work.

\section{Methodology}

In this section, we introduce the basic structure of standard PINNs and the learning principles underlying the proposed PIEFL framework.

\subsection{Standard PINNs}

In this paper, we mainly consider PDEs in the following general form:
\begin{equation}
	\begin{aligned}
		\frac{\partial u(x,t)}{\partial t}&+\mathcal{N}\left[u(x,t)\right]=0,\quad x\in\Omega\subset\mathbb{R}^D,\ t\in[0,T],\\\
		u(x,0)&=I(x),\quad x\in\Omega,\\\
		u(x,t)&=B(x,t),\quad x\in\partial\Omega.
	\end{aligned}
	\label{eq:1}
\end{equation}
Here, $u(x,t)$ denotes the target solution, and $\mathcal{N}\left[u(x,t)\right]$ represents a nonlinear differential operator. $I(x)$ and $B(x,t)$ denote the initial condition and boundary condition (IC/BC), respectively. According to Ref. \cite{raissi2019physics}, the solution $u(x,t)$ can be approximated by a deep neural network. In essence, PINNs incorporate physical laws into the loss function through physical constraints. Therefore, the physical residual $f$ corresponding to Eq. (1) is defined as
\begin{equation}
	f:=\frac{\partial u}{\partial t}+\mathcal{N}[u].
	\label{eq:2}
\end{equation}

The loss function of PINNs consists of both the data loss $MSE_{u}$ and the physical loss $MSE_{f}$:
\begin{equation}
	MSE=MSE_{u}+MSE_{f},
	\label{eq:3}
\end{equation}
where
\begin{equation}
	MSE_{u}=\frac{1}{N_u}\sum_{i=1}^{N_u}{|u(x_u^i,t_u^i)-u^i|^2},
	\label{eq:4}
\end{equation}
and
\begin{equation}
	MSE_{f}=\frac{1}{N_f}\sum_{i=1}^{N_f}{|f(x_f^i,t_f^i)|^2}.
	\label{eq:5}
\end{equation}
Here, $(t_u^i,x_u^i,u^i), i=1,\ldots,N_u$ denote the initial and boundary training data, while ${(t_f^i,x_f^i)}, i=1,\ldots,N_f$ represent the collocation points used for evaluating the physical residual $f$. The term $MSE_{u}$ corresponds to the loss associated with the initial and boundary conditions, whereas $MSE_{f}$ enforces the physical constraint defined by Eq. (2) on the collocation point set.

\subsection{Physics-Informed Error Field Learning Framework}

In this section, we provide a detailed description of the fundamental principles of the Physics-Informed Error Field Learning (PIEFL) framework. PIEFL adopts a novel cascaded architecture that decomposes the solution procedure into two stages. In the first stage, a primary network is trained to obtain an initial approximate solution satisfying the governing PDE. In the second stage, an error network, guided by the prediction of the primary network, is introduced to specifically learn the error field subject to the corresponding physical constraints. Finally, the high-accuracy solution is obtained by combining the outputs of the two networks. The overall architecture of the proposed framework is illustrated in Fig. 1.

\begin{figure}[h!]%
	\centering
	\includegraphics[width=0.85\textwidth]{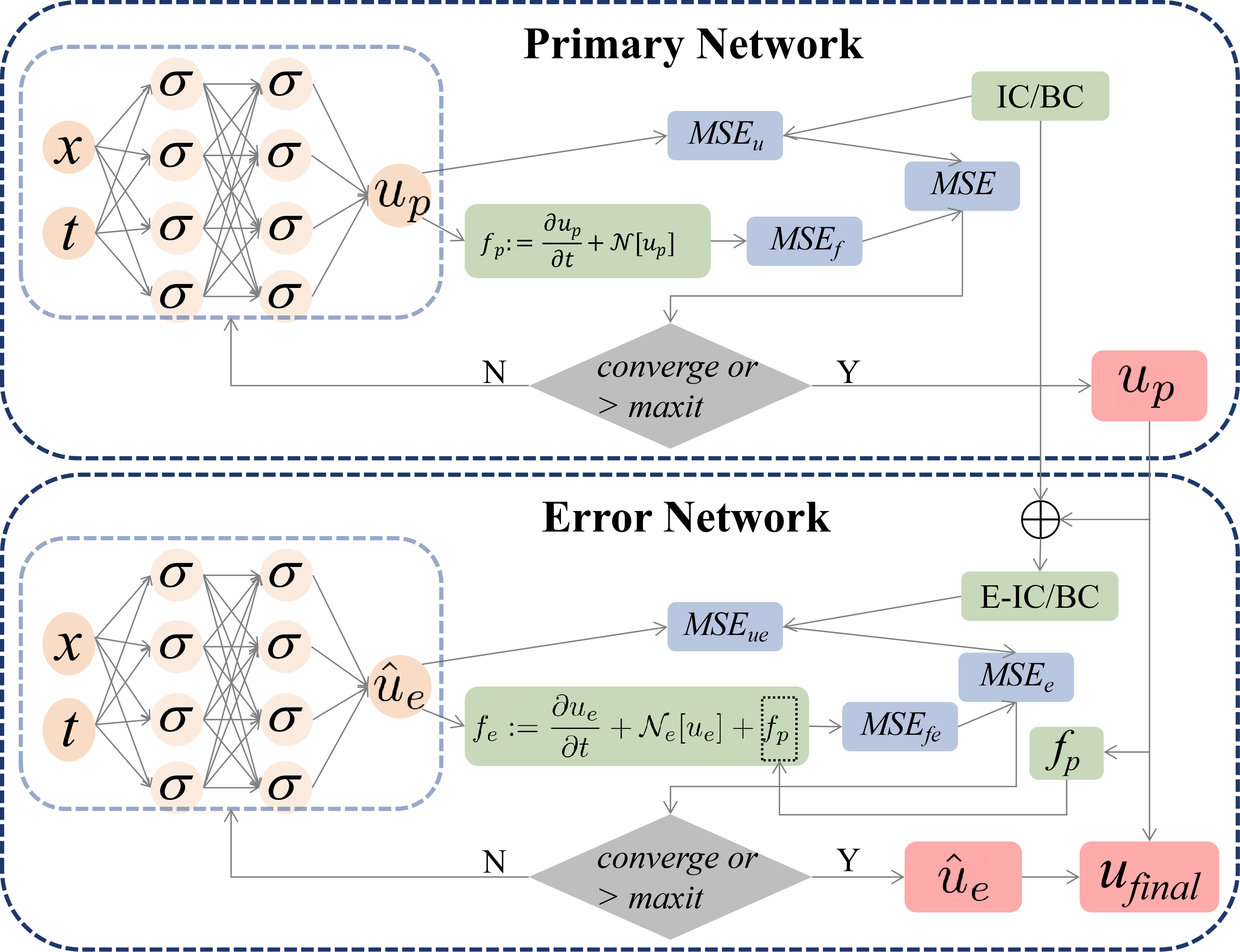}
	\caption{Schematic diagram of PIEFL }\label{fig1}
\end{figure}

\subsubsection{Primary Network}

In the PIEFL framework, the primary network performs the first-stage solution task, aiming to provide a fundamental approximate solution to the target PDE that satisfies both the governing physical laws and the corresponding IC/BC constraints.

The primary network shares the same input configuration and loss function as standard PINNs. Its output consists of the initial prediction $u_p$ and the corresponding PDE residual:
\begin{equation}
	f_p:=\frac{\partial u_p}{\partial t}+\mathcal{N}[u_p].
	\label{eq:6}
\end{equation}

The predicted solution $u_p$ and the residual $f_p$ are subsequently involved in the training of the error network, providing the initial predictive information and physical residual constraints required for subsequent error correction.
\subsubsection{Error Network}

The error network is the core component for implementing subsequent error correction within the PIEFL framework. After the parameters of the primary network are fixed, the basic approximate solution $u_p$ generated by the primary network is used as the preliminary solution information. The objective of the error network is to learn and reconstruct a physically constrained error field governed by the corresponding physical equations. 
In this paper, the error field is defined as $u_e$, which satisfies
\begin{equation}
	u_e=u-u_p.
	\label{eq:7}
\end{equation}

By rewriting Eq. (7) as $u=u_p+u_e$ and substituting it into Eq. (1), we obtain
\begin{equation}
	\frac{\partial (u_p+u_e)}{\partial t}+\mathcal{N}[u_p+u_e]=0.
	\label{eq:8}
\end{equation}
Substituting Eq. (6) into Eq. (8) yields
\begin{equation}
	f_p-\mathcal{N}[u_p]+\frac{\partial u_e}{\partial t}+\mathcal{N}[u_p+u_e]=0.
	\label{eq:9}
\end{equation}
Here, the nonlinear differential operator associated with the error field is defined as
\begin{equation}
	\mathcal{N}_e[u_e]=\mathcal{N}[u_p+u_e]-\mathcal{N}[u_p].
	\label{eq:10}
\end{equation}

Therefore, the governing equation of the error field can be expressed as
\begin{equation}
	\frac{\partial u_e}{\partial t}+\mathcal{N}_e[u_e]+f_p=0.
	\label{eq:11}
\end{equation}
Accordingly, the corresponding physical residual of the error network is defined as
\begin{equation}
	f_e:=\frac{\partial u_e}{\partial t}+\mathcal{N}_e[u_e]+f_p.
	\label{eq:12}
\end{equation}

Based on the physical residual $f_e$, the loss function of the error network in PIEFL is constructed as
\begin{equation}
	MSE_e=MSE_{ue}+MSE_{fe},
	\label{eq:13}
\end{equation}
where
\begin{equation}
	MSE_{ue}=\frac{1}{N_{ue}}\sum_{i=1}^{N_{ue}}|u_e(x_u^i,t_u^i)-u_e^i|^2,
	\label{eq:14}
\end{equation}
and
\begin{equation}
	MSE_{fe}=\frac{1}{N_{fe}}\sum_{i=1}^{N_{fe}}|f_e(x_f^i,t_f^i)|^2.
	\label{eq:15}
\end{equation}

Similar to standard PINNs, $\{(t_u^i,x_u^i,u_e^i)\}_{i=1}^{N_{ue}}$ denote the initial and boundary training data for the error field, while $\{(t_f^i,x_f^i)\}_{i=1}^{N_{fe}}$ represent the collocation points used for evaluating $f_e$. It should be noted that the error initial and boundary conditions (E-IC/BC) are obtained by subtracting the primary network prediction $u_p$ from the original IC/BC.

After training, the final prediction of PIEFL at any spatiotemporal coordinate $(x,t)$ is given by
\begin{equation}
	u_{\mathrm{final}}(x,t)=u_p(x,t)+\hat{u}_e(x,t).
	\label{eq:16}
\end{equation}
Here, $\hat{u}_e(x,t)$ denotes the output of the error network. Since the magnitude of the error field $u_e(x,t)$ is typically small, an output scaling coefficient $\alpha$ is introduced to amplify the error network output and reduce the difficulty of learning the error field.

\section{Numerical Experiments}

In this section, several numerical experiments are conducted to compare the performance of standard PINNs and PIEFL. Specifically, three representative PDEs are considered, including the KdV equation, the nonlinear Schrödinger equation, and the (2+1)-dimensional Kadomtsev–Petviashvili (KP) equation. The relative $\mathbb{L}_{2}$ error is employed as the evaluation metric to quantify the performance of different models. All implementations are developed using Python 3.7 and TensorFlow 1.15.
\subsection{KdV Equation}

The KdV equation is given by
\begin{equation}
	\begin{aligned}
		u_t+6uu_x+u_{xxx}=0.
	\end{aligned}
	\label{eq:13}
\end{equation}

In this subsection, the one-soliton solution is considered as the target solution \cite{Moloney1989new}:
\begin{equation}
	\begin{aligned}
		u=8a^2\frac{\theta}{(1+\theta)^2},
	\end{aligned}
	\label{eq:14}
\end{equation}
where $\theta=ax-4a^2t+\sigma$, and $a$ and $\sigma$ are arbitrary constants.

For the primary network, the training target is the solution field $u$. The corresponding physical residual is defined as
\begin{equation}
	\begin{aligned}
		f_p(x,t):=u_t+6uu_x+u_{xxx}.
	\end{aligned}
\end{equation}

For the error network, the training target is the error field $u_e$. Substituting $u=u_p+u_e$ into Eq. (17) yields
\begin{equation}
	\begin{aligned}
		\frac{\partial u_p}{\partial t}
		+6u_p\frac{\partial u_p}{\partial x}
		+\frac{\partial^3u_p}{\partial x^3}
		+\frac{\partial u_e}{\partial t}
		+6u_p\frac{\partial u_e}{\partial x}
		+6u_e\frac{\partial u_p}{\partial x}
		+6u_e\frac{\partial u_e}{\partial x}
		+\frac{\partial^3u_e}{\partial x^3}=0.
	\end{aligned}
\end{equation}

By substituting Eq. (19) into Eq. (20), the governing equation of the error field is obtained as
\begin{equation}
	\begin{aligned}
		&\frac{\partial u_e}{\partial t}
		+6u_p\frac{\partial u_e}{\partial x}
		+6u_e\frac{\partial u_p}{\partial x}
		+6u_e\frac{\partial u_e}{\partial x}
		+\frac{\partial^3u_e}{\partial x^3}
		+f_p(x,t)=0.
	\end{aligned}
\end{equation}

Accordingly, the physical residual of the error network is defined as
\begin{equation}
	\begin{aligned}
		f_e(x,t):=
		\frac{\partial u_e}{\partial t}
		+6u_p\frac{\partial u_e}{\partial x}
		+6u_e\frac{\partial u_p}{\partial x}
		+6u_e\frac{\partial u_e}{\partial x}
		+\frac{\partial^3u_e}{\partial x^3}
		+f_p(x,t).
	\end{aligned}
\end{equation}

The solution domain is set as $[-10,10]\times[-1,1]$. The numbers of training points and collocation points are set to 200 and 8000, respectively. The network architectures of both the primary network and the error network are set to $[2,20,20,20,20,1]$. In PIEFL, the Adam optimization iterations for the primary network and the error network are set to 5,000 and 10,000, respectively. To evaluate the effectiveness of the error network, a standard PINN model trained for 15,000 iterations is employed as the baseline comparison. The scaling coefficient $\alpha$ of the error network is set to 0.001.

\begin{figure}
	\centering
	\includegraphics[width=0.6\textwidth]{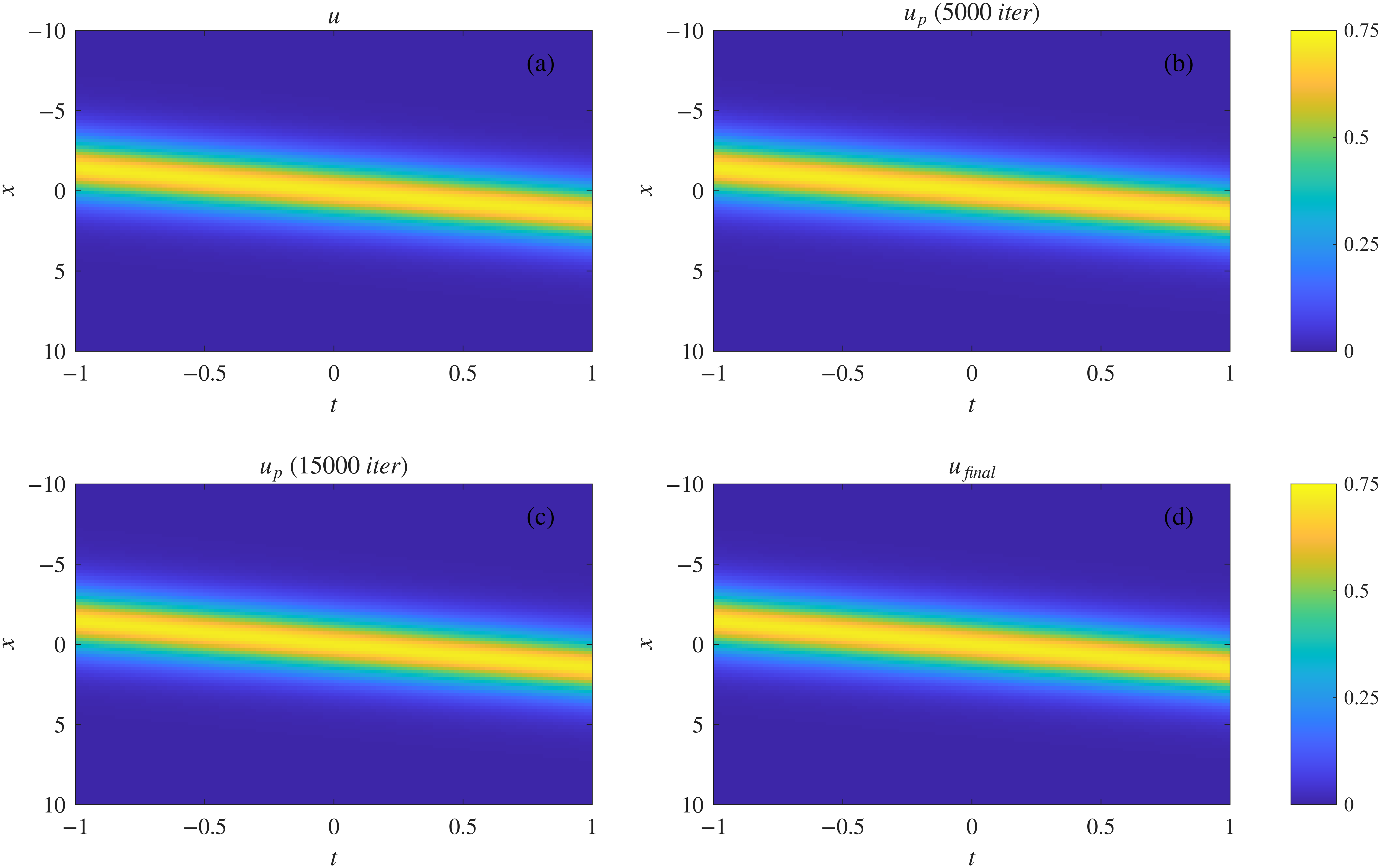}
	\caption{Solutions of the KdV equation: (a) Exact solution; (b) Prediction of the primary network; (c) Prediction of the standard PINN model; (d) Prediction of the PIEFL model.}\label{fig1}
\end{figure}

After training, the relative error of the primary network prediction $u_p$ is $3.846063\mathrm{e}{-3}$, while the relative error of the error network prediction $\hat{u}_{e}$ is $1.659345\mathrm{e}{-2}$. The relative error of the final prediction is reduced to $1.469116\mathrm{e}{-4}$. In comparison, the baseline PINN model achieves a relative error of $9.480554\mathrm{e}{-4}$. Fig. 2 illustrates the predicted solutions obtained by standard PINNs and PIEFL. Fig. 3 presents the corresponding error fields and correction fields.

As shown in Fig. 2, both the standard PINN model and the PIEFL model are capable of accurately solving the KdV equation. Fig. 3 demonstrates that the error field generated by the standard PINN model exhibits clear spatial-temporal patterns and a strong correlation with the target solution. Meanwhile, the proposed error network effectively learns the error field and provides accurate corrections to the primary network prediction. Furthermore, Fig. 4 illustrates the evolution of the relative errors of PINNs, PIEFL, and the error network, thereby revealing the relationship between the prediction accuracy of the error field and the overall solution accuracy. It should be noted that both PIEFL and the error network are based on the prediction results of PINNs that have been trained for 5,000 iterations. Therefore, the PIEFL and error network curves in Fig. 4 both start at 5,000 iterations, while the PINNs curve shows the evolution of the error as the primary network continues training up to 15,000 iterations.

\begin{figure}
	\centering
	\includegraphics[width=0.6\textwidth]{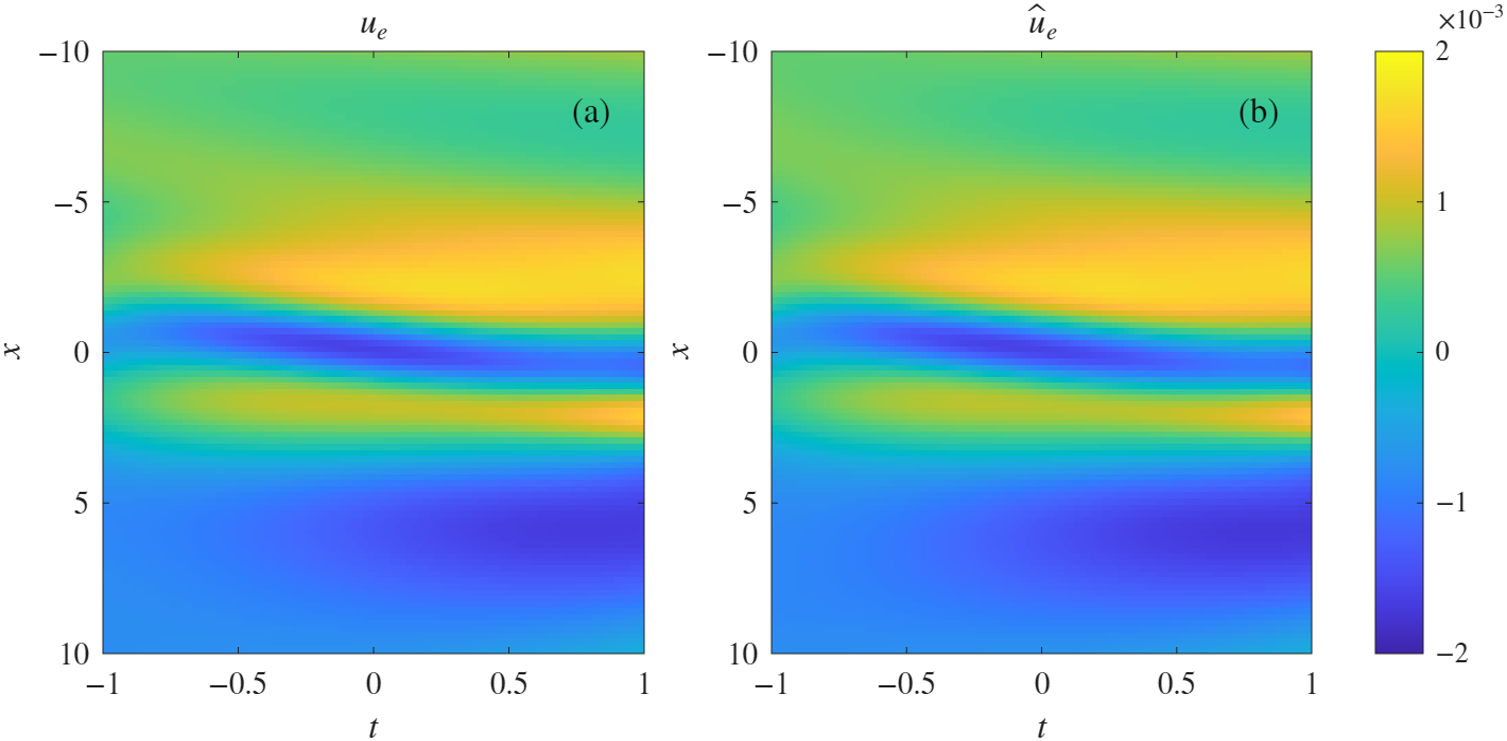}
	\caption{Error fields of the KdV equation: (a) Error field $u_e$; (b) Correction field $\hat{u}_e$.}\label{fig1}
\end{figure}
\begin{figure}
	\centering
	\includegraphics[width=0.75\textwidth]{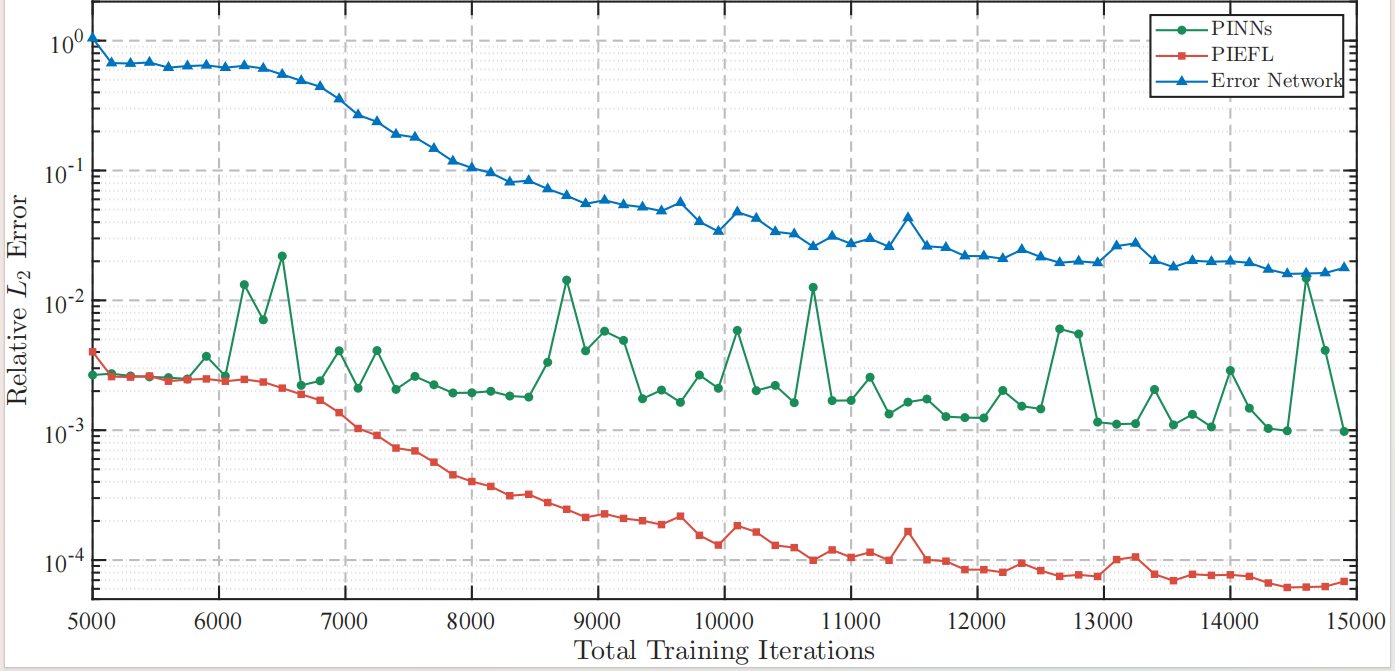}
	\caption{Evolution of the relative errors of PINNs, PIEFL, and the error network during the late-stage training.}\label{fig1}
\end{figure}

From the comparison between PINNs and PIEFL, it can be seen that in the initial stage, PIEFL’s total error was slightly higher than that of PINNs. Subsequently, the two exhibited relatively similar trends in error changes during the early stages. This indicated that the error network has not yet sufficiently learned the remaining error in the PINNs prediction. As the training proceeds, the PIEFL error begins to decrease more rapidly at approximately 6500 iterations. Remarkably, at around 7000 iterations, PIEFL already achieves a level of accuracy comparable to the final result obtained by PINNs after 15000 iterations. In contrast, although the error of PINNs continued to decrease during subsequent training, the rate of improvement slowed, and its error curve exhibited more pronounced fluctuations. These results indicate that, during the later stages of PINNs training, the accuracy gains from continuously optimizing the original network follow a diminishing trend, whereas PIEFL can more efficiently improve the accuracy of the solution by directly learning from the residual errors in existing predictions. 

From the relationship between the error network’s error and the total PIEFL error, the two curves generally exhibit similar patterns of change. As the prediction accuracy of the error network continues to improve, the overall error of PIEFL decreases in tandem, indicating a direct link between the quality of error field learning and the improvement in the final solution’s accuracy. In particular, when the error in the error network decreases significantly, the PIEFL error typically decreases accordingly. This result demonstrates that the improvement in PIEFL’s final accuracy primarily stems from the error network’s effective learning of the residual errors from the primary network. Even if the error network cannot completely and precisely reconstruct the entire error field, its effective capture of the primary error features can still translate into a further improvement in the accuracy of the overall solution.

A series of experiments is further conducted to investigate the necessity of introducing the output scaling coefficient $\alpha$ and its influence on the solution accuracy. The values of $\alpha$ are set to 1, 0.1, 0.01, 0.001, and 0.0001, and the corresponding results are presented in Table 1. To quantitatively evaluate the effect of $\alpha$, the error reduction rate (ERR) is introduced:
\begin{equation}
	\begin{aligned}
		ERR=\frac{Error_1-Error_2}{Error_1},
	\end{aligned}
\end{equation}
where $Error_1$ denotes the prediction error of the PINN model and $Error_2$ represents the prediction error of PIEFL. In Table 1, $Error_1$ corresponds to the prediction error of the standard PINN model after 15,000 iterations.

\begin{table}[width=.9\linewidth,cols=4,pos=h]
\caption{KdV equation: Relative $\mathbb{L}_{2}$ errors between the predicted and exact solutions for different scaling coefficients}
\begin{tabular*}{\tblwidth}{@{} Lccccc@{} }
\toprule
$\alpha$ & 1 & 0.1 & 0.01&0.001&0.0001\\
\midrule
Error($\hat{u}_{e}$) & $7.483748\mathrm{e}{-1}$ & $6.965753\mathrm{e}{-1}$ & $4.498170\mathrm{e}{-2}$ &  $3.737533\mathrm{e}{-2}$ & $3.805477\mathrm{e}{-2}$  \\
Error($u_{final}$)  & $2.878297\mathrm{e}{-3}$ & $2.679072\mathrm{e}{-3}$ & $1.730028\mathrm{e}{-4}$& $1.437481\mathrm{e}{-4}$& $1.463611\mathrm{e}{-4}$ \\
ERR  &  $-203.60{\%}$& $-182.58{\%}$ &  81.75${\%}$& 84.83${\%}$& 84.56${\%}$ \\
\bottomrule
\end{tabular*}%
\end{table}

As shown in Table 1, the value of $\alpha$ has a significant influence on the performance of the error network. When $\alpha$ is set to 1 or 0.1, the output magnitude of the error network is not properly scaled, and the network fails to achieve sufficient prediction accuracy within 10,000 iterations due to the difficulty of learning the small-amplitude error field. When $\alpha$ is set to 0.01, 0.001, or 0.0001, the output scaling effectively enlarges the magnitude of the error network prediction, allowing the network to better capture the characteristics of the error field and provide more effective corrections to the primary network prediction.
\subsection{Nonlinear Schrödinger Equation}

The nonlinear Schrödinger equation is given by
\begin{equation}
	\begin{aligned}
		iq_t+q_{xx}+2|q|^2q=0.
	\end{aligned}
	\label{eq:16}
\end{equation}
Here, $q$ is a complex-valued function, which can be decomposed as $q=u+iv$, where $u(x,t)$ and $v(x,t)$ are real-valued functions. Therefore, Eq. (16) can be transformed into the following coupled real-valued system:
\begin{equation}
	\begin{aligned}
		u_t+v_{xx}+2(u^2+v^2)v&=0,\\
		v_t-u_{xx}-2(u^2+v^2)u&=0.
	\end{aligned}
	\label{eq:17}
\end{equation}

In this subsection, both one-soliton and two-soliton solutions are considered as target solutions. According to Ref. \cite{yang2010nonlinear}, the $N$-soliton solution is expressed as
\begin{equation}
	q=-2i\frac{\det F}{\det M},
	\label{eq:18}
\end{equation}
where $M$ is an $N\times N$ matrix,
\begin{equation}
	M=
	\left[
	\begin{array}{cccc}
		M_{11}&M_{12}&\cdots&M_{1N}\\
		M_{21}&M_{22}&\cdots&M_{2N}\\
		\vdots&\vdots&\vdots&\vdots\\
		M_{N1}&M_{N2}&\cdots&M_{NN}
	\end{array}
	\right],
	\label{eq:19}
\end{equation}
and $F$ is an $(N+1)\times(N+1)$ matrix,
\begin{equation}
	F=
	\left[
	\begin{array}{cccc}
		0&c_1e^{\theta_1}&\cdots&c_Ne^{\theta_N}\\
		e^{-\theta_1^*}&M_{11}&\cdots&M_{1N}\\
		\vdots&\vdots&\vdots&\vdots\\
		e^{-\theta_N^*}&M_{N1}&\cdots&M_{NN}
	\end{array}
	\right].
	\label{eq:20}
\end{equation}
Here,
\[
M_{jk}=\frac{e^{-(\theta_k+\theta_j^*)}+c_j^*c_ke^{\theta_k+\theta_j^*}}
{\zeta_j^*-\zeta_k},
\]
\[
\theta_k=-i\zeta_kx-2i\zeta_k^2t,\quad (j,k=1,2,\cdots,N),
\]
and $\zeta_k$ and $c_k$ $(j,k=1,2,\cdots,N)$ are complex constants. When $N=1$ and $N=2$, the corresponding one-soliton and two-soliton solutions can be obtained.

For the one-soliton solution, we set $c_1=1$ and $\zeta_1=0.1+0.3i$. The solution domain is defined as $[-10,10]\times[0,1]$. For the two-soliton solution, we set $c_1=-1/2$, $c_2=1/2$, $\zeta_1=0.1+0.3i$, and $\zeta_2=0.1+0.2i$. The solution domain is defined as $[-15,15]\times[-2,2]$.

For the primary network, the training targets are the solution components $u$ and $v$. The corresponding physical residuals are defined as
\begin{equation}
	\begin{aligned}
		f_p^u(x,t)&:=u_t+v_{xx}+2(u^2+v^2)v,\\
		f_p^v(x,t)&:=-v_t+u_{xx}+2(u^2+v^2)u.
	\end{aligned}
\end{equation}

For the error network, the training targets are the error components $u_e$ and $v_e$. Substituting $u=u_p+u_e$ and $v=v_p+v_e$ into Eq. (17), we obtain
\begin{equation}
	\begin{aligned}
		&\frac{\partial u_p}{\partial t}
		+\frac{\partial v_p}{\partial x}
		+2(u_p^2+v_p^2)v_p
		+\frac{\partial u_e}{\partial t}
		+\frac{\partial v_e}{\partial x}
		+2\lambda_1v_e
		+2\lambda_2(v_p+v_e)=0,\\
		&-\frac{\partial v_p}{\partial t}
		+\frac{\partial u_p}{\partial x}
		+2(u_p^2+v_p^2)u_p
		-\frac{\partial v_e}{\partial t}
		+\frac{\partial u_e}{\partial x}
		+2\lambda_1u_e
		+2\lambda_2(u_p+u_e)=0,
	\end{aligned}
\end{equation}
where $\lambda_1=u_p^2+v_p^2$ and $\lambda_2=u_e^2+v_e^2+2u_pu_e+2v_pv_e$.

By substituting Eq. (29) into Eq. (30), the governing equations of the error field become
\begin{equation}
	\begin{aligned}
		&\frac{\partial u_e}{\partial t}
		+\frac{\partial v_e}{\partial x}
		+2\lambda_1v_e
		+2\lambda_2(v_p+v_e)
		+f_p^u(x,t)=0,\\
		&-\frac{\partial v_e}{\partial t}
		+\frac{\partial u_e}{\partial x}
		+2\lambda_1u_e
		+2\lambda_2(u_p+u_e)
		+f_p^v(x,t)=0.
	\end{aligned}
\end{equation}
Accordingly, the corresponding physical residuals are defined as
\begin{equation}
	\begin{aligned}
		f_e^u(x,t)&:=\frac{\partial u_e}{\partial t}
		+\frac{\partial v_e}{\partial x}
		+2\lambda_1v_e
		+2\lambda_2(v_p+v_e)
		+f_p^u(x,t),\\
		f_e^v(x,t)&:=-\frac{\partial v_e}{\partial t}
		+\frac{\partial u_e}{\partial x}
		+2\lambda_1u_e
		+2\lambda_2(u_p+u_e)
		+f_p^v(x,t).
	\end{aligned}
\end{equation}

The network architectures of both the primary network and the error network are set to $[2,50,50,50,50,50,2]$. The numbers of training points and collocation points are set to 600 and 10000, respectively. In PIEFL, the Adam optimization iterations for both the primary network and the error network are set to 5,000. To evaluate the effectiveness of the error network, a standard PINN model trained for 10,000 iterations is employed as the baseline comparison. The scaling coefficient $\alpha$ of the error network is set to 0.001. The experimental results are presented in Table 2.
\begin{table}[width=.9\linewidth,cols=4,pos=h]
\caption{Nonlinear Schrödinger equation: Relative $\mathbb{L}_{2}$ errors of one-soliton and two-soliton solutions}
\begin{tabular*}{\tblwidth}{@{} Lccccc@{} }
\toprule
solution & primary network & error network &PIEFL& PINNs(10000 \it{iter}) &ERR\\
\midrule
one-soliton solution& $2.773287\mathrm{e}{-3}$ &$2.105211\mathrm{e}{-3}$  & $8.745380\mathrm{e}{-5}$ &  $1.829887\mathrm{e}{-3}$ & $95.22\%$ \\
two-soliton solution & $1.086328\mathrm{e}{-2}$ & $2.592568\mathrm{e}{-2}$ & $3.573394\mathrm{e}{-4}$& $4.199116\mathrm{e}{-3}$&  $91.49\%$\\
			\hline
\end{tabular*}
\end{table}

The relative $\mathbb{L}_{2}$ errors of the primary network are $2.773287\mathrm{e}{-3}$ and $1.086328\mathrm{e}{-2}$ for the one-soliton and two-soliton solutions, respectively. After error correction by the error network, these errors are reduced to $8.745380\mathrm{e}{-5}$ and $3.573394\mathrm{e}{-4}$. Compared with the standard PINN model, the corresponding ERRs are $95.22\%$ and $91.49\%$, respectively. Although the standard PINN model is capable of accurately solving both cases, the proposed error network provides effective correction to the primary network prediction and achieves substantial improvements in solution accuracy.
\begin{figure}
	\centering
	\includegraphics[width=0.6\textwidth]{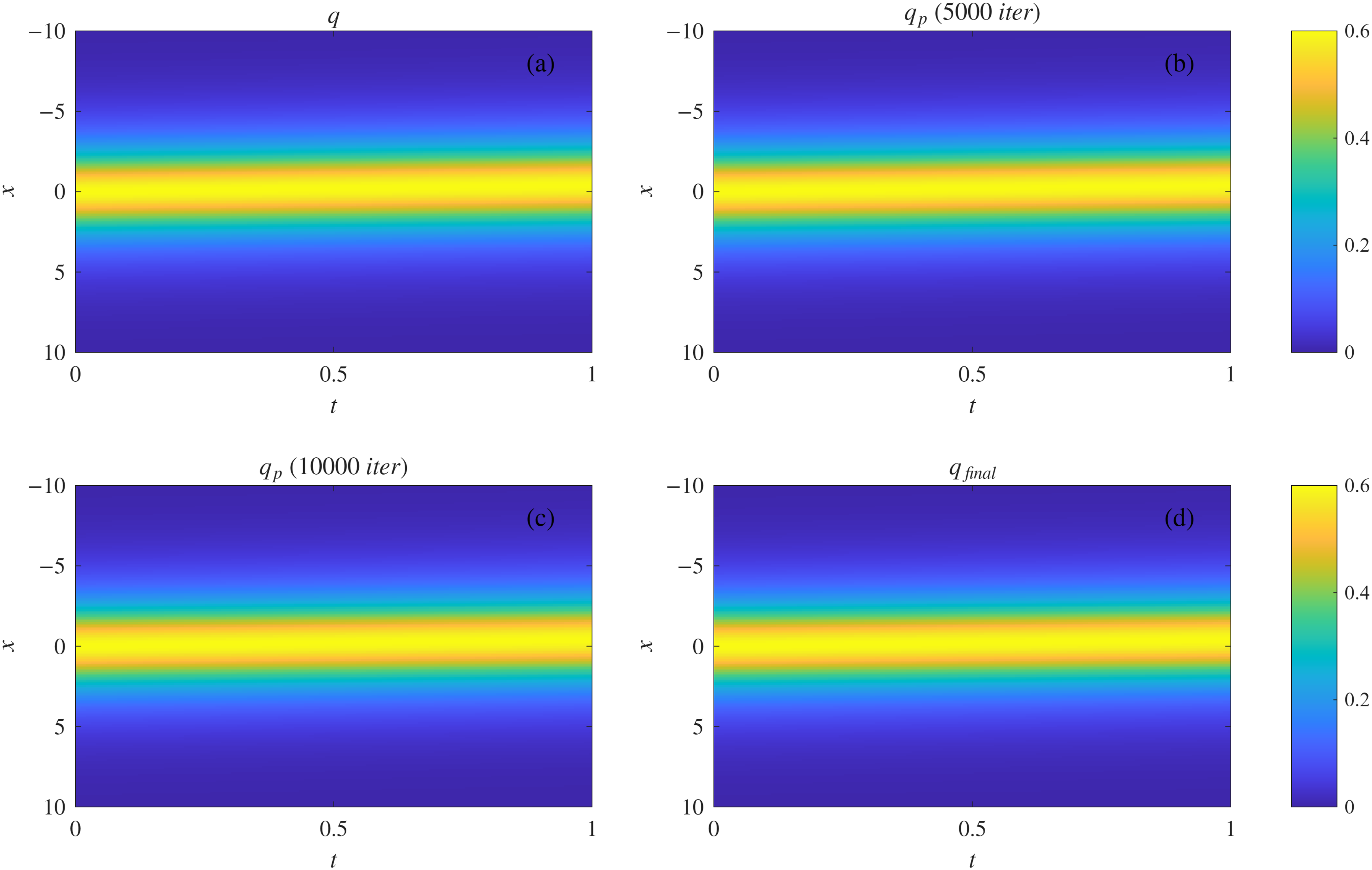}
	\caption{The one-soliton solution of the nonlinear Schrödinger equation: (a) Exact solution; (b) Prediction of the primary network; (c) Prediction of the standard PINN model; (d) Prediction of the PIEFL model.}\label{fig1}
\end{figure}
\begin{figure}
	\centering
	\includegraphics[width=0.6\textwidth]{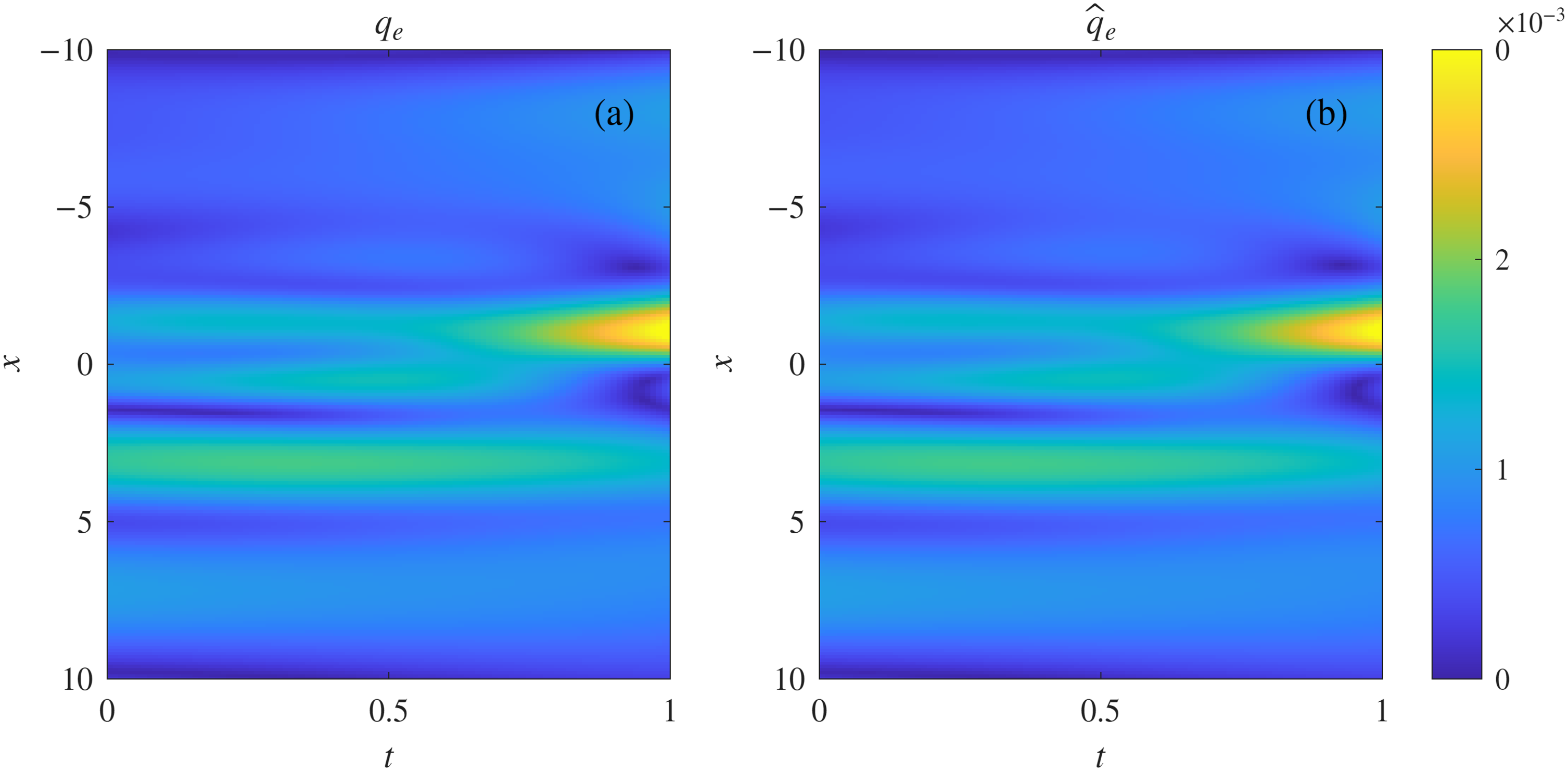}
	\caption{Error fields of the one-soliton solution of the nonlinear Schrödinger equation: (a) Error field $q_e$; (b) Correction field $\hat{q}_e$.}\label{fig1}
\end{figure}
\begin{figure}
	\centering
	\includegraphics[width=0.6\textwidth]{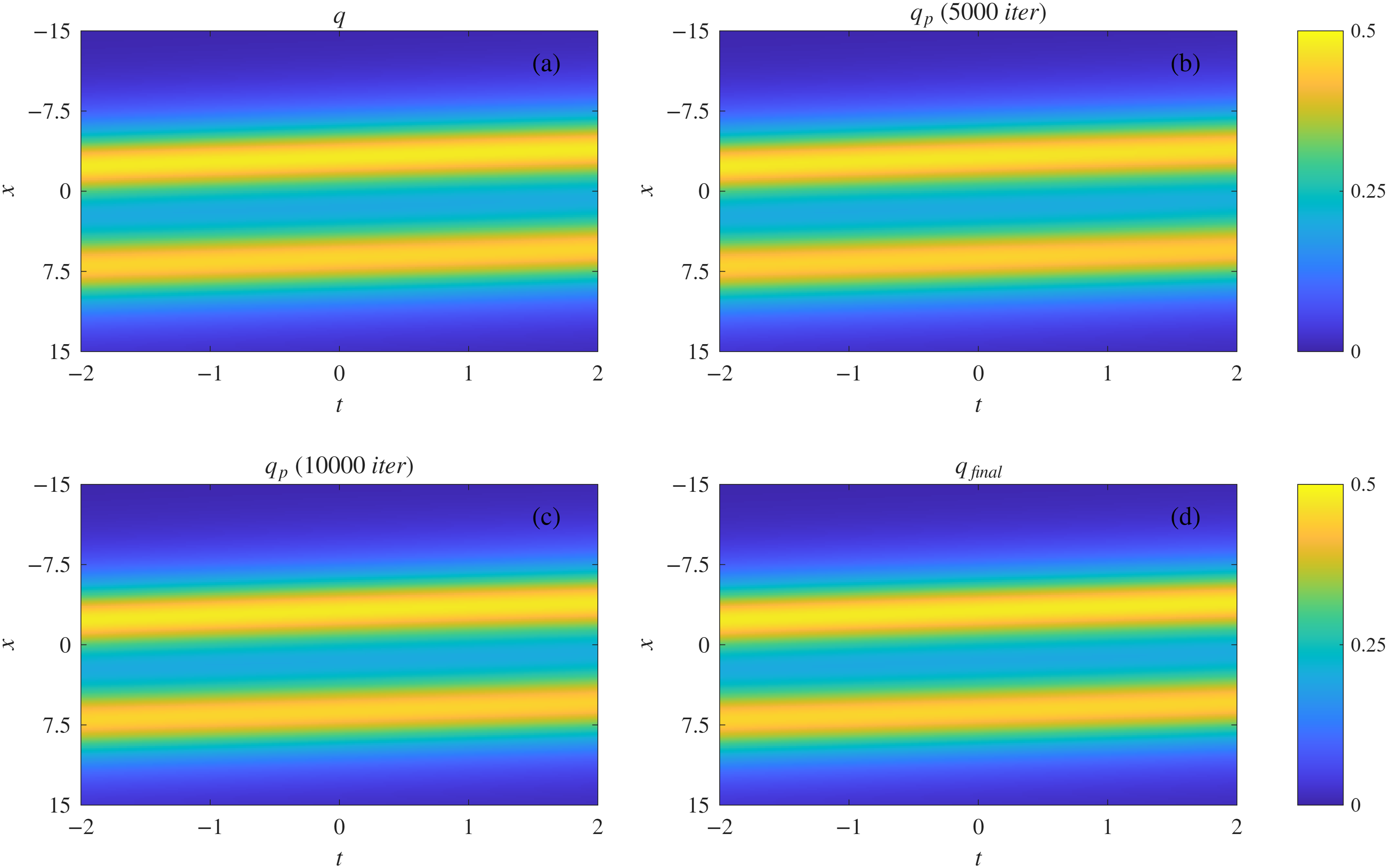}
	\caption{The two-soliton solution of the nonlinear Schrödinger equation: (a) Exact solution; (b) Prediction of the primary network; (c) Prediction of the standard PINN model; (d) Prediction of the PIEFL model.}\label{fig1}
\end{figure}

\begin{figure}
	\centering
	\includegraphics[width=0.6\textwidth]{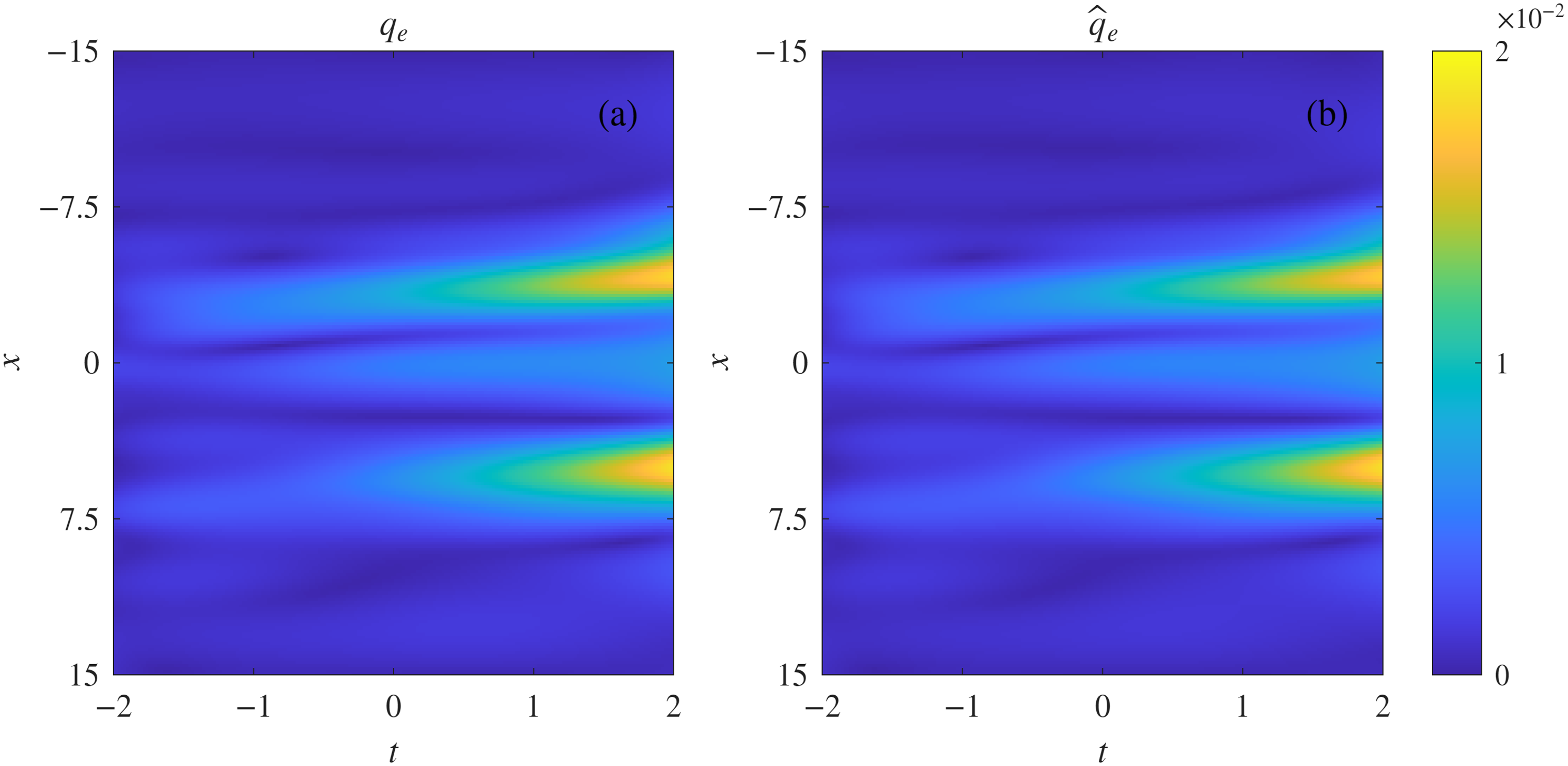}
	\caption{Error fields of the two-soliton solution of the nonlinear Schrödinger equation: (a) Error field $q_e$; (b) Correction field $\hat{q}_e$.}\label{fig1}
\end{figure}

Fig. 5 and Fig. 7 present the spatiotemporal distributions of the one-soliton and two-soliton solutions corresponding to the cases reported in Table 2, respectively. Fig. 6 and Fig. 8 illustrate the corresponding error distributions. Consistent with the results in Table 2, both PINNs and PIEFL are capable of accurately solving these two cases. Furthermore, PIEFL effectively learns the error field associated with the primary PINN prediction, thereby providing accurate error corrections and further improving the solution accuracy. 
Overall, the proposed error correction strategy effectively enhances the performance of PINNs for solving the nonlinear Schrödinger equation.

\subsection{(2+1)-Dimensional Kadomtsev--Petviashvili Equation}

The (2+1)-dimensional KP equation is given by
\begin{equation}
	\begin{aligned}
		(u_t+u_{xxx}+6uu_x)_x+\sigma^2u_{yy}=0.
	\end{aligned}
	\label{eq:21}
\end{equation}

In this subsection, the one-soliton and two-soliton solutions are considered as target solutions \cite{miao2022physics}:
\begin{equation}
	\begin{aligned}
		u_1(x,y,t)=2[\ln(1+\exp(\eta_1))]_{xx},
	\end{aligned}
	\label{eq:22}
\end{equation}
\begin{equation}
	\begin{aligned}
		u_2(x,y,t)=2[\ln(1+\exp(\eta_1)+\exp(\eta_2)
		+\exp(\eta_1+\eta_2+A_{12}))]_{xx}.
	\end{aligned}
	\label{eq:23}
\end{equation}
where
\[
\eta_i=k_i[x+p_iy-(k_i^2+\sigma^2p_i^2)t],\quad (i=1,2),
\]
and
\[
\exp(A_{12})=
\frac{3(k_1-k_2)^2-\sigma^2(p_1-p_2)^2}
{3(k_1+k_2)^2-\sigma^2(p_1-p_2)^2}.
\]

For the one-soliton solution, the initial and boundary conditions are obtained by setting $k_1=1$, $p_1=2$, and $\sigma=1$. The spatial domain and temporal interval are defined as $\Omega=[-5,5]\times[-5,5]$ and $[T_1,T_2]=[-2,2]$, respectively. Similarly, for the two-soliton solution, we set $k_1=1$, $k_2=4/5$, $p_1=1$, $p_2=4/5$, and $\sigma=1$, with $\Omega=[-10,10]\times[-10,10]$ and $[T_1,T_2]=[-2,2]$.

For the primary network, the training target is the solution field $u$. The corresponding physical residual is defined as
\begin{equation}
	\begin{aligned}
		f_p(x,y,t):=(u_t+u_{xxx}+6uu_x)_x+\sigma^2u_{yy}.
	\end{aligned}
\end{equation}

For the error network, the training target is the error field $u_e$. The corresponding physical residual is defined as
\begin{equation}
	\begin{aligned}
		f_e(x,y,t):={}
		\frac{\partial}{\partial x}
		\left(
		\frac{\partial u_e}{\partial t}
		+\frac{\partial^3u_e}{\partial x^3}
		+6u_p\frac{\partial u_e}{\partial x}
		+6u_e\frac{\partial u_p}{\partial x}
		+6u_e\frac{\partial u_e}{\partial x}
		\right)
		+\sigma^2\frac{\partial^2u_e}{\partial y^2}
		+f_p(x,y,t).
	\end{aligned}
\end{equation}

The numbers of training points and collocation points are set to 500 and 10000, respectively. The network architectures of both the primary network and the error network are set to $[3,50,50,50,50,1]$. The scaling coefficient $\alpha$ of the error network is set to 0.001.

For the one-soliton solution, the Adam optimization iterations of the primary network and the error network in PIEFL are set to 3,000 and 2,000, respectively. In the comparison experiment, the standard PINN model is trained for 5,000 iterations. For the two-soliton solution, the Adam optimization iterations of both the primary network and the error network in PIEFL are set to 5,000. In the comparison experiment, the standard PINN model is trained for 10,000 iterations. The experimental results are presented in Table 3.

\begin{table}[width=.9\linewidth,cols=4,pos=h]
\caption{ KP equation: Relative $\mathbb{L}_{2}$ errors of one-soliton and two-soliton solutions}
\begin{tabular*}{\tblwidth}{@{} Lccccc@{} }
\toprule
solution & primary network & error network & PIEFL & PINNs &ERR\\
\midrule
one-soliton solution& $1.254043\mathrm{e}{-2}$ &$7.678907\mathrm{e}{-2}$  & $9.629841\mathrm{e}{-4}$ &  $5.039722\mathrm{e}{-3}$ & $80.89\%$ \\
two-soliton solution & $1.493969\mathrm{e}{-2}$ & $3.923684\mathrm{e}{-1}$ & $5.861866\mathrm{e}{-3}$& $9.5369942\mathrm{e}{-3}$&  38.53$\%$\\
\bottomrule
\end{tabular*}%
\end{table}

For the one-soliton solution, the primary network achieves a relative $\mathbb{L}_{2}$ error of $1.254043\mathrm{e}{-2}$. After correction by the learned error field, this error is further reduced to $9.629841\mathrm{e}{-4}$. Compared with the standard PINN model, PIEFL reduces the relative error by 80.89\%, demonstrating the effectiveness of error field learning.
For the two-soliton solution, although the improvement achieved by error field learning is less pronounced, PIEFL still exhibits a significant advantage over the standard PINN model, achieving an ERR of 38.53\%. Fig. 9 and Fig. 11 present the spatiotemporal distributions of the one-soliton and two-soliton solutions corresponding to the cases reported in Table 3, respectively. Fig. 10 and Fig. 12 illustrate the corresponding error distributions.
This indicates that, for more complex nonlinear solution structures, the error field remains informative, although its correction capability may depend on the complexity of the underlying solution field.

\begin{figure}[htbp]
	\centering
	\includegraphics[width=0.7\textwidth]{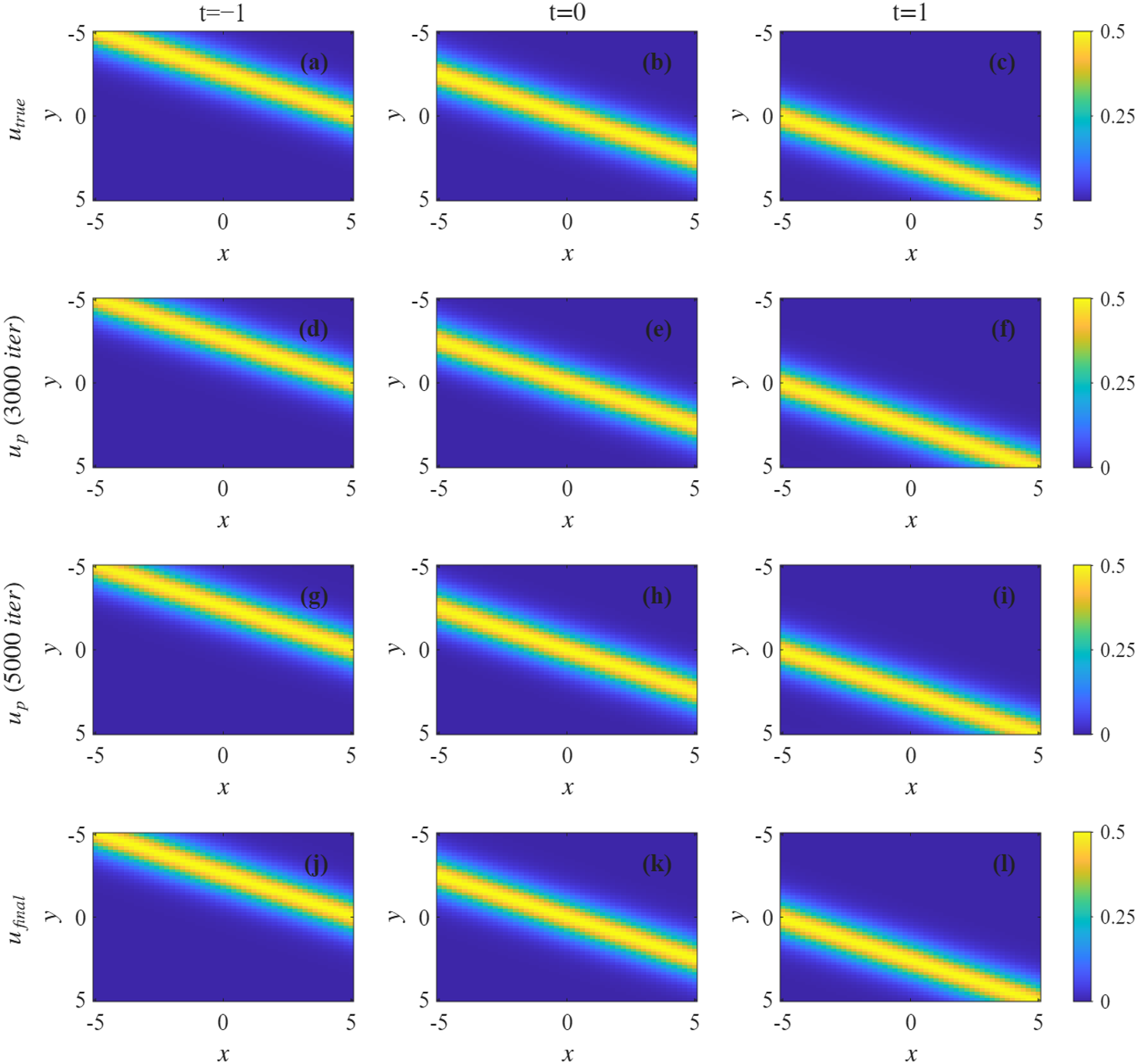}
	\caption{The one-soliton solution of the KP equation: (a) Exact solution; (b) Predicted solution of the primary network; (c) Predicted solution of the PINNs; (d) Predicted solution of the PIEFL }\label{fig1}
\end{figure}
\begin{figure}[htbp]
	\centering
	\includegraphics[width=0.7\textwidth]{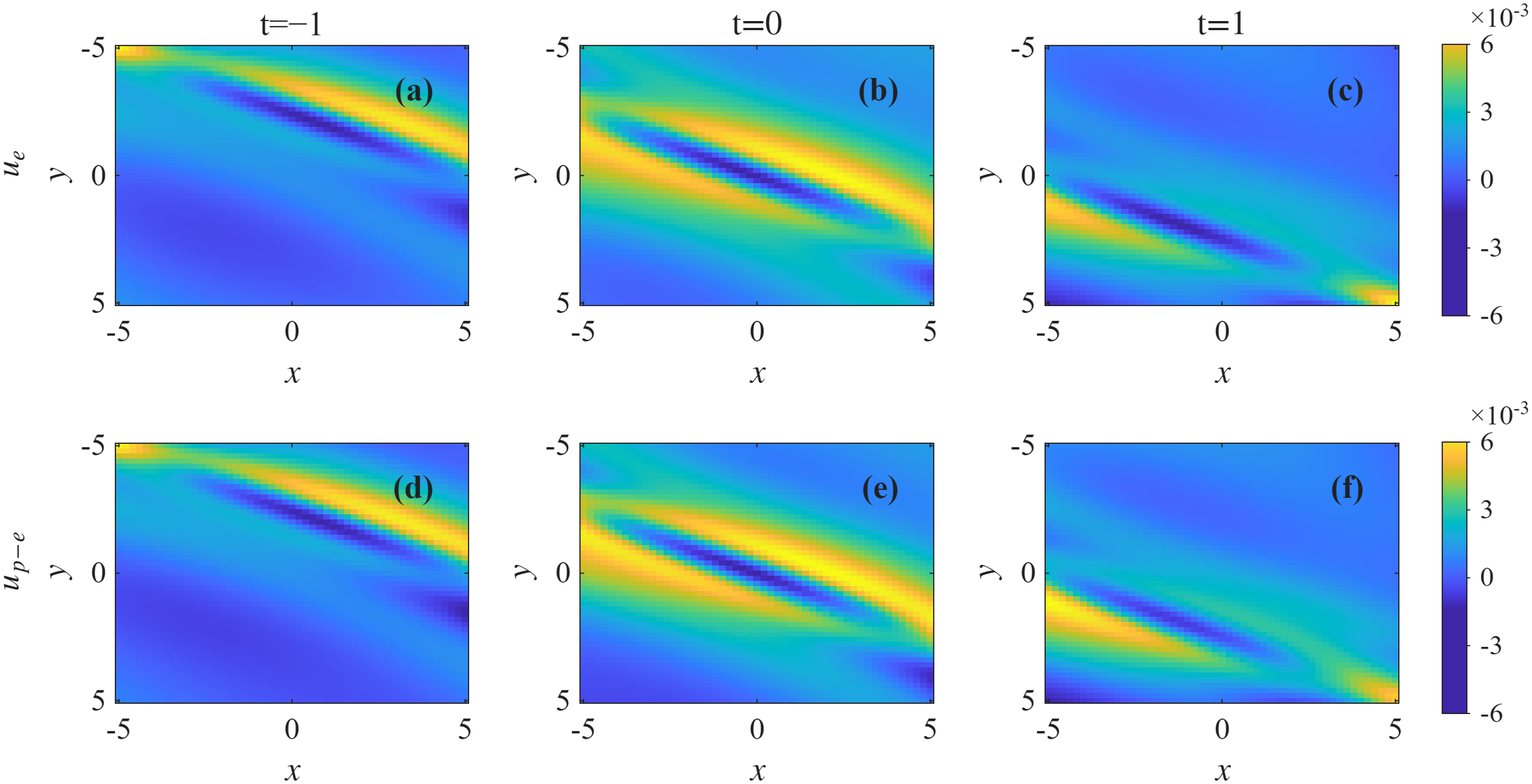}
	\caption{The one-soliton solution of the KP equation: (a) The error field $u_e$; (b) The correction field $\hat{u}_{e}$ }\label{fig1}
\end{figure}

For the one-soliton solution, the primary network captures the overall propagation behavior of the solitary wave; however, non-negligible deviations remain in the predicted solution field. After applying PIEFL, the prediction results are further refined through error field correction. The visualization of the corresponding error field indicates that the prediction error exhibits clear spatial structures rather than random fluctuations, suggesting that the error field contains valuable information that can be effectively learned for subsequent correction.

\begin{figure}[htbp]
	\centering
	\includegraphics[width=0.7\textwidth]{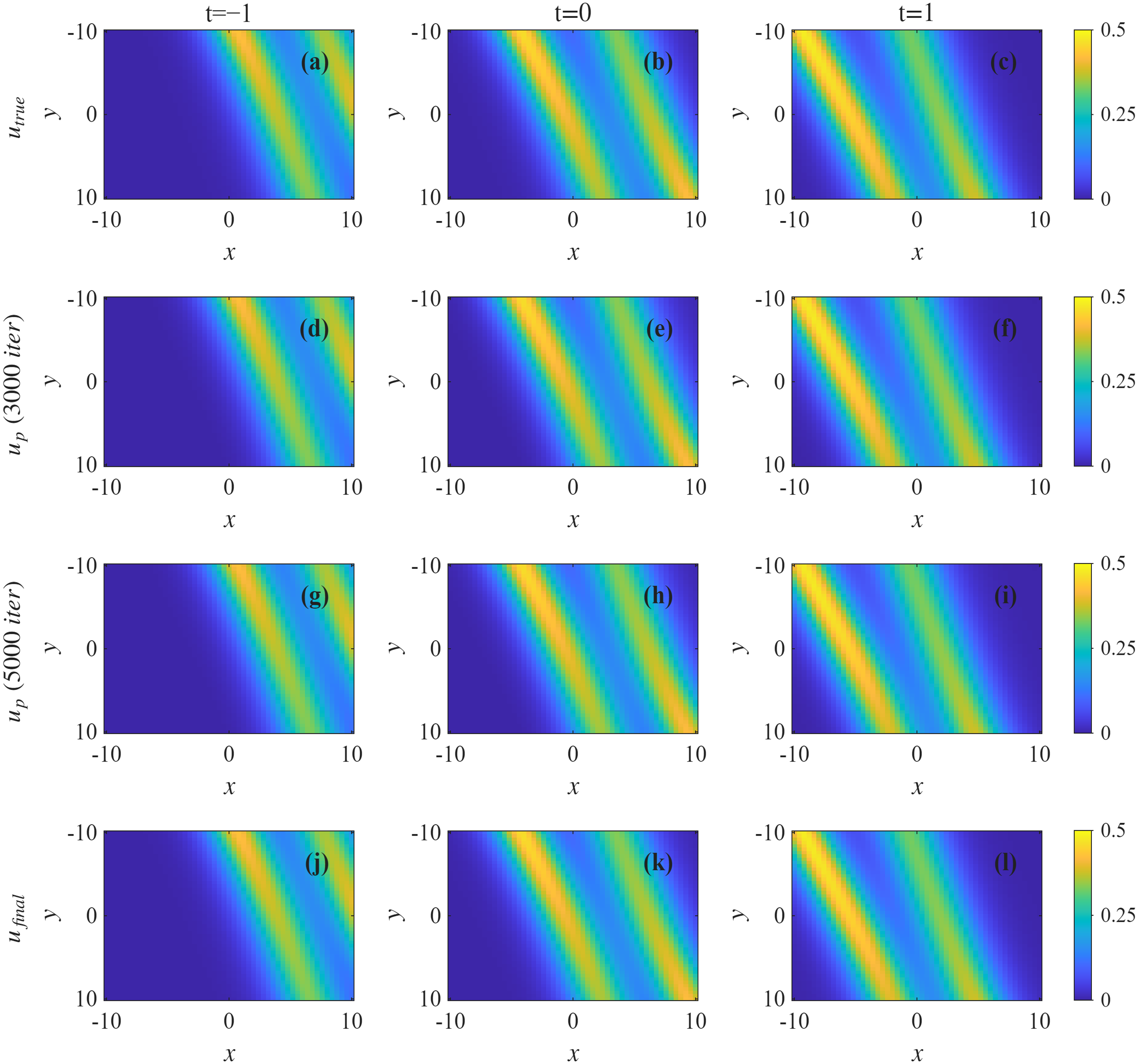}
	\caption{The two-soliton solution of the KP equation: (a) Exact solution; (b) Predicted solution of the primary network; (c) Predicted solution of the PINNs; (d) Predicted solution of the PIEFL }\label{fig1}
\end{figure}

\begin{figure}[htbp]
	\centering
	\includegraphics[width=0.7\textwidth]{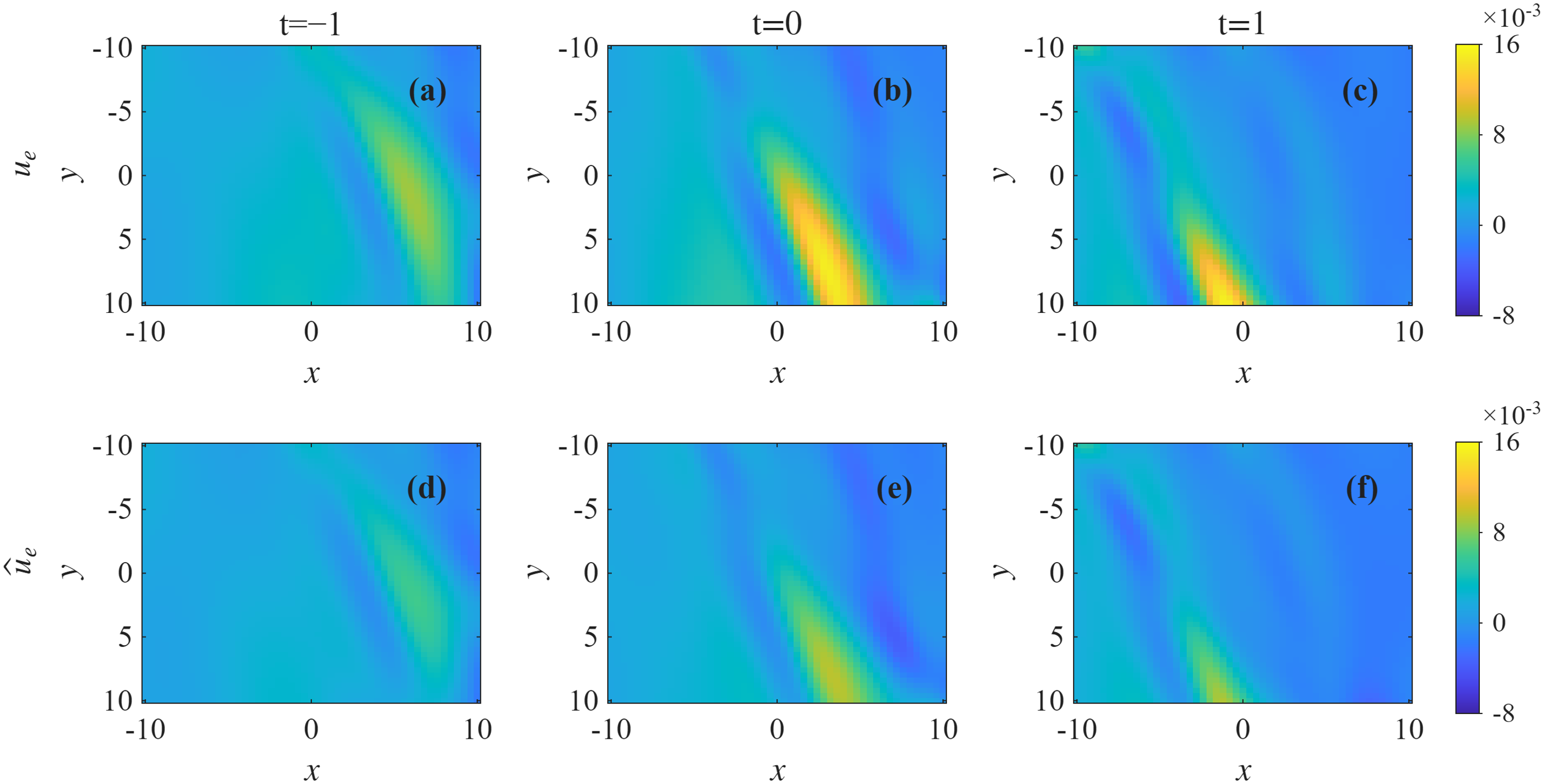}
	\caption{The two-soliton solution of the KP equation: (a) The error field $u_e$; (b) The correction field $\hat{u}_{e}$ }\label{fig1}
\end{figure}

For the two-soliton solution, learning the error field becomes more challenging due to the increased complexity of the solution structure. As shown in Fig. 12, the learned error field cannot fully recover the exact error distribution, indicating that accurately representing the error field is more difficult for complex solutions. Nevertheless, PIEFL still achieves a significant improvement in the final prediction accuracy, reducing the error from $1.493969\mathrm{e}{-2}$ of the primary network to $5.861866\mathrm{e}{-3}$. These results demonstrate that, even when PIEFL cannot precisely reconstruct the entire error field, the learned error information can still provide effective correction, thereby reducing prediction errors and improving the accuracy of the final solution.

\section{Conclusion}
This paper proposes a Physics-Informed Error Field Learning (PIEFL) framework to improve the computational efficiency of PINNs during the late-stage optimization process. Unlike conventional PINNs, which continuously approximate the complete solution field using a single network, PIEFL introduces an independent error network after the primary network achieves high accuracy, thereby shifting the learning objective from the solution field to the error field. By deriving the governing equations associated with the error field, the error learning process is incorporated into the physics-informed optimization framework, enabling the network to learn prediction errors while preserving the physical constraints of the original governing equations.

The effectiveness of the proposed PIEFL framework is validated through several representative PDE problems, including the KdV equation, the nonlinear Schrödinger equation, and the KP equation. Numerical results demonstrate that, compared with continuous training strategies of conventional PINNs, PIEFL can further reduce prediction errors with additional error correction. For solution fields with relatively simple structures, the error network effectively captures the dominant features of the error field and achieves accurate corrections. For more complex solution structures, although accurately approximating the error field becomes more challenging, PIEFL can still capture useful error information and improve the accuracy of the final solution.

By modifying the learning objective rather than continuously increasing the optimization cost of the original solution network, the proposed method provides a new perspective for improving the computational efficiency of PINNs. Since PIEFL does not require modifications to existing PINN architectures, it can be seamlessly integrated with existing optimization strategies, including network architecture enhancement, adaptive sampling, and loss balancing techniques. Future work will focus on developing more efficient error field learning strategies and extending their applications to complex physical systems and high-dimensional PDEs.

\section*{Conflict of Interests}
The authors declare that there is no conflict of interests regarding the publication of this paper.

\printcredits

\bibliographystyle{cas-model2-names}

\bibliography{reference.bib}



\end{document}